\documentclass{article}
\usepackage[preprint]{flexible}

\usepackage[utf8]{inputenc} % allow utf-8 input
\usepackage[T1]{fontenc}    % use 8-bit T1 fonts
\usepackage{hyperref}       % hyperlinks
\usepackage{url}            % simple URL typesetting
\usepackage{booktabs}       % professional-quality tables
\usepackage{amsfonts}       % blackboard math symbols
\usepackage{nicefrac}       % compact symbols for 1/2, etc.
\usepackage{microtype}      % microtypography
\usepackage{xcolor}         % colors
\usepackage{graphicx}
\usepackage{caption} % 加载caption来使用\captionof
\usepackage{amsmath}
\usepackage{fontawesome5}
\usepackage{natbib}
\setcitestyle{numbers,square}
\DeclareMathOperator*{\Concat}{\Vert}

\title{Flexible ViG: Learning the Self-Saliency for Flexible Object Recognition}

\author{%
  \textbf{Lin Zuo, Kunshan Yang, Xianlong Tian,} \\ 
  \textbf{Kunbin He, Yongqi Ding, Mengmeng Jing\textsuperscript{*}} \\
  University of Electronic Science and Technology of China \\
  linzuo@uestc.edu.cn, \\
  \{jingmeng1992, tianxianlong3, hekunbin19\}@gmail.com \\
  \{ksyang, yqding\}@std.uestc.edu.cn \\
}

\begin{document}

\maketitle
\begin{abstract}
Existing computer vision methods mainly focus on the recognition of rigid objects, whereas the recognition of flexible objects remains unexplored. Recognizing flexible objects poses significant challenges due to their inherently diverse shapes and sizes, translucent attributes, ambiguous boundaries, and subtle inter-class differences. In this paper, we claim that these problems primarily arise from the lack of object saliency. To this end, we propose the Flexible Vision Graph Neural Network (FViG) to optimize the self-saliency and thereby improve the discrimination of the representations for flexible objects. Specifically, on one hand, we propose to maximize the channel-aware saliency by extracting the weight of neighboring nodes, which adapts to the shape and size variations in flexible objects. On the other hand, we maximize the spatial-aware saliency based on clustering to aggregate neighborhood information for the centroid nodes, which introduces local context information for the representation learning. To verify the performance of flexible objects recognition thoroughly, for the first time we propose the Flexible Dataset (FDA), which consists of various images of flexible objects collected from real-world scenarios or online. Extensive experiments evaluated on our Flexible Dataset demonstrate the effectiveness of our method on enhancing the discrimination of flexible objects.
\end{abstract}

\section{Introduction}
Computer vision \cite{he2016deep, vaswani2017attention, carion2020end, dosovitskiy2020image} has been widely employed in various applications, ranging from simple object recognition to complex scene understanding \cite{miikkulainen2024evolving, mehrani2023self}. Image recognition is categorized into two main types: rigid and non-rigid objects recognition. Rigid objects, as illustrated in Figure 1 (a), maintain consistent shapes and sizes regardless of their positioning or viewing angle. In the past decades, the recognition of rigid objects has been well explored and shown excellent performance due to their invariant geometrical structure and appearance property. Notably, the convolutional neural network could effectively extract the features of the rigid objects \cite{he2016deep}. Non-rigid objects, however, exhibit inconsistent shapes or sizes, resulting in various appearances depending on their positions and viewing angles. Their variations in shape and size pose challenges in recognition. Flexible objects, a subset of non-rigid objects, show even larger variations in size and shape. As illustrated in Figure 1 (b), flexible objects, e.g., clouds, smoke, water, flames, and glare, not only vary greatly in shape and size but may be semi-transparent and lack clear boundaries \cite{jha2022point}, which makes it extremely difficult to extract discriminative representations. Furthermore, the small inter-class differences among flexible objects make them challenging to distinguish.

Recognizing flexible objects is critical in numerous application fields. For example, recognizing flames and smoke is vital for early fire detection \cite{wang2020rapid}, precise recognition of different cloud formations plays a key role in weather forecasting and climate monitoring \cite{zhang2018cloudnet}, and accurate recognition of elements such as glare \cite{wu2021how} is essential for achieving realism and visual coherence in composite imagery. Despite the importance of these tasks, a fundamental requirement for training effective
\begin{minipage}[t]{0.48\textwidth}
\vspace{0pt} % 保证内容从顶部开始
discriminative deep models is the availability of high-quality datasets. However, existing data sets dedicated to fire \cite{jadon2019firenet, foggia2015real} and cloud \cite{krizhevsky2009learning, zhang2018cloudnet} are narrowly focused on singular recognition tasks, thus demonstrating limited challenges and diversities. These datasets do not include various flexible objects, leading to models trained on them exhibiting diminished discriminative capabilities. To solve this problem, we propose the FDA dataset, specifically designed for flexible objects research. FDA comprises a diverse range of flexible objects images gathered from real-world scenarios and online resources, providing extensive training and testing data.
\end{minipage}%
\hfill %
\begin{minipage}[t]{0.50\textwidth} % Use half of the text width
  \centering
  \vspace{0pt} % 保证内容从顶部开始
  \includegraphics[width=\linewidth]{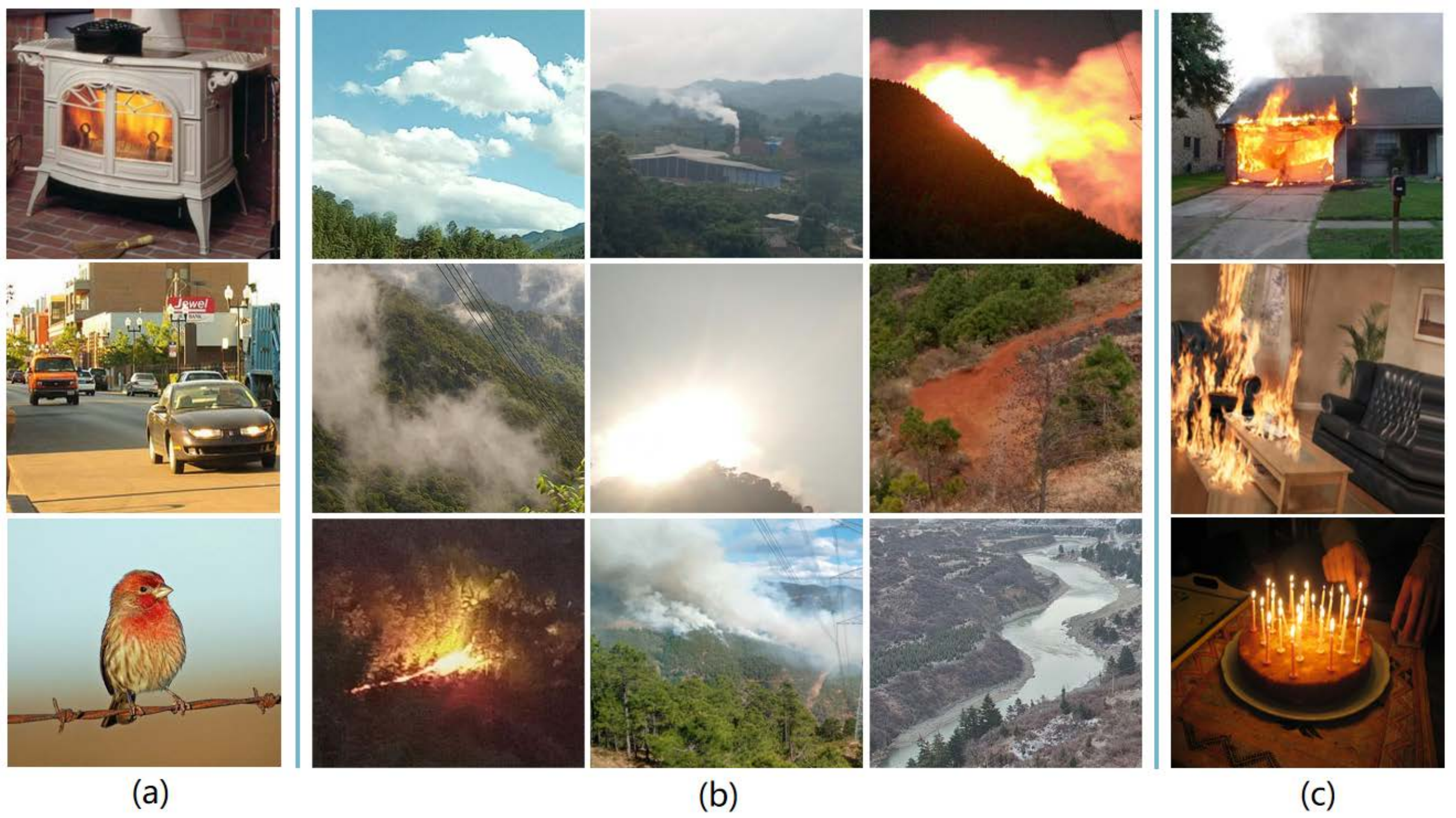}
  \captionof{figure}{(a) rigid objects. (b) flexible objects images from our proposed FDA. (c) fire images from FireNet dataset.}
\end{minipage}

Although several studies \cite{jadon2019firenet, wang2020rapid} have attempted to recognize flexible objects, they typically focus on recognizing one or two specific types within images, not addressing the need for detailed differentiation and recognition of multiple categories of flexible objects. Two primary issues impede the recognition of flexible objects: firstly, the diversity and irregularity in their shapes and sizes, especially their translucent physical properties or unclear boundaries, hinder the extraction of consistent representational features; secondly, the small inter-class differences make them hard to distinguish. These issues stem from a lack of saliency in flexible objects. Saliency refers to the distinct and prominent features that make objects stand out from their surroundings and differentiate them from other categories of objects.

To mitigate these problems, we propose Flexible ViG in this paper, which is primarily optimized for saliency. Channel-aware saliency is employed to address the difficulty in extracting consistent representation caused by the diverse variations in the shape and size of flexible objects. Spatial-aware saliency is utilized to address the challenge of identifying flexible objects with minimal inter-class differences. By optimizing both channel and spatial aspects, Flexible ViG enhances the discrimination capabilities for flexible objects. Extensive experiments demonstrate the effectiveness of our FViG method in capturing neighboring relationship and enhancing the representation capacity of central nodes. The main contributions of this work are as follows.

(1) We propose Flexible ViG, designed to maximize self-saliency. In which, channel-aware saliency adaptively captures neighboring relationships by extracting channel weight information, matching the various shapes and sizes of flexible objects. This improves the model’s sensitivity to fine-grained features, enhancing its ability to distinguish flexible objects from backgrounds. Spatial-aware saliency enhances node representation through node-level clustering, aggregating overlooked neighboring node information to update central nodes.

(2) We have created a dataset named FDA, which consists of diverse images of flexible objects collected from real-world scenarios or online sources. To the best of our knowledge, FDA is the first extensive, multi-category dataset specifically designed for the recognition of flexible objects. This dataset establishes a benchmark for evaluating the performance of models in flexible objects recognition tasks.

(3) Extensive experiments on the FDA dataset demonstrate the effectiveness of our proposed FViG, achieving recognition performance comparable to other state-of-the-art methods. Furthermore, these experiments validate that the FDA dataset serves as a reliable benchmark to evaluate the performance of various methods.

\section{Related Work}

\subsection{Graph Neural Network}
Graph Neural Networks (GNNs) \cite{pedronette2014unsupervised, waikhom2021graph}, Graph Convolutional Networks (GCNs) \cite{zhang2019graph, kipf2016semi}, and Graph Attention Networks (GATs) \cite{velickovic2017graph, gu2021discovering} have received significant attention in recent years for their ability to handle data with complex relationships and topological structures. These models excel in analyzing graph-structured data and are applied in diverse fields, including social networks, recommendation systems, and molecular chemistry. GNNs \cite{waikhom2021graph} focus on the acquisition of node representations through dynamic exchange and aggregation of information between nodes. The advent of GCNs \cite{kipf2016semi} introduced the concept of convolution to structured non-Euclidean spaces. There are two primary types: spectral-based, which leverages Fourier transforms and graph signal processing through eigen decomposition; and spatial-based, which defines convolutions directly on the graph, updating the feature representations of central nodes via a message-passing mechanism. Velickovic et al. \cite{velickovic2017graph} introduced the GAT model, which incorporates attention mechanisms into GNN to dynamically weight the influence of neighboring nodes during the aggregation process, focusing on more significant nodes. However, these methods are primarily suited for structured data, such as in text processing, and are less effective for learning features with unstructured data such as images. In contrast, our FViG is capable of adapting to these unstructured images.

\subsection{Graph Vision Model}
Shen et al. \cite{shen2023git} introduced the interactive graph transformer (GiT), designed for vehicle re-identification. Zheng et al. \cite{yun2022graph} developed a graph transformer network designed for whole-slide image representations, incorporating a novel transformer fusion method. Furthering the combination of graph convolution and self-attention, Lin et al. \cite{lin2021mesh} presented Mesh Graphormer, a technique for reconstructing human poses and meshes from single images. Gu \cite{gu2021discovering} introduced GANR, a network representation based on graph attention, which employs the attention mechanism to uncover and quantify the relationships and significance of the nodes, demonstrating superior performance in applications such as link prediction. Ma \cite{ma2021graph} explored the local structural details of the graphs through GAT and introduced GAT-POS, an enhancement to GAT that incorporates positional embeddings to represent the structural and positional data of the nodes. GAT-POS recorded remarkable results on heterogeneous graph-structured datasets. Marking a pioneering fusion of graph structures with images, Han et al. \cite{han2022vision} proposed the Vision Graph Neural Network (ViG). This network treats each image patch as an individual graph node and utilizes a k-NN approach to establish relationships between these patches. Despite its innovative approach, ViG primarily captures image patch similarity, potentially overlooking the latent manifold structure of the image. Furthermore, the construction of graph structures is crucial, and Li \cite{li2019deepgcns} emphasized the importance of studying different distance metrics to build graph structures. However, ViG adopts the Euclidean distance to measure the dependencies between nodes for graph construction. This method could not adequately capture the complex geometric relationships between different patches \cite{lv2018color, ling2007shape, lomov2020skeleton, kim2023vit, li2019deepgcns, pedronette2014unsupervised}. In contrast, we propose channel-aware saliency learning to adaptively capture neighboring relationships and learn fine-grained graph representations.

\subsection{Clustering}
Clustering constitutes an unsupervised learning algorithm characterized by the generation of clusters, each comprising a collection of data objects. Within these clusters, objects display pronounced intracluster similarity and markedly reduced intercluster similarity. A quintessential example of such algorithms is K-means, introduced by MacQueen \cite{macqueen1967some}. This method seeks to determine k centroids within a dataset and to partition the dataset into k distinct clusters, with the objective of minimizing the sum of squared distances from each data point to its nearest centroid. CLARA \cite{schubert2019faster} strategically extracts a subset of data points as a representative sample from the dataset, subsequently employing the Partitioning Around Medoids (PAM) algorithm on this sample to improve computational efficiency. Furthermore, Affinity propagation clustering, pioneered by Frey et al. \cite{frey2007clustering} is recognized as a rapid and effective clustering algorithm.

Both Vision GNN \cite{han2022vision} and deepGCN \cite{li2019deepgcns} employ dilated graph convolutions, which significantly expand the receptive field, address oversmoothing issues during model training, and improve graph representation capabilities. However, these methods do not fully utilize the information from the nodes in dilated regions. Such neglected nodes are critical for graph representation learning. Inspired by Ma's proposed method of patch-level aggregation and dispatch \cite{ma2023image}, we propose spatial-aware saliency learning based on node-level feature clustering to effectively establish graph learning relationships among nodes within dilated regions, thereby enhancing the overall model's learning and representation capacity.

\section{Method}
Saliency refers to the distinct and prominent features that enable objects to stand out from their surroundings and facilitate differentiation from other object categories. Subsequently, as shown in Figure 2, we optimize saliency from both channel-aware and spatial-aware aspects.

\begin{figure*}[h]
  \centering
  \includegraphics[width=0.94\linewidth]{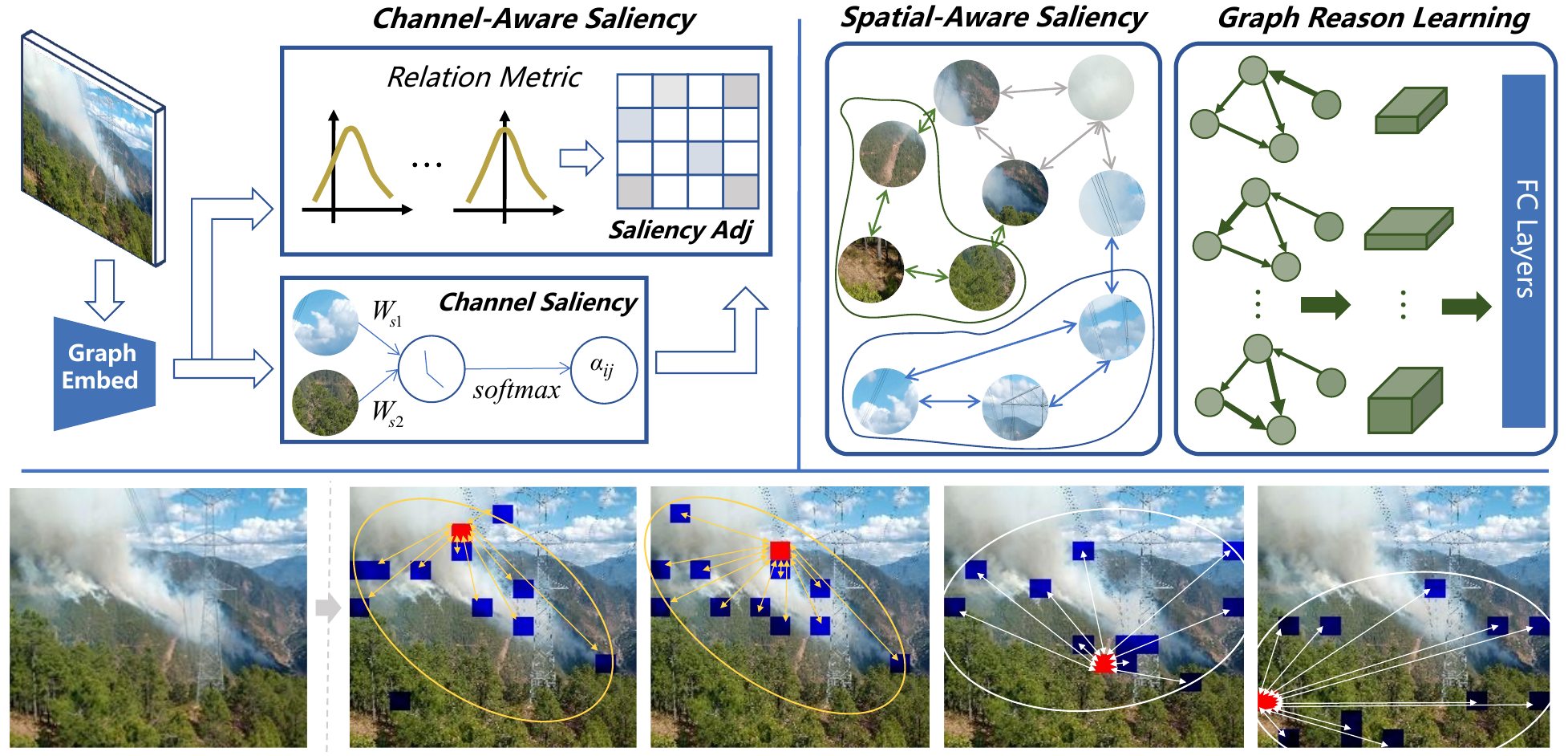}
  \caption{The top section describes the workflow of the proposed FViG, encompassing graph embedding, relation metrics, graph attention, graph generation and clustering, and graph reasoning learning. The bottom section details the graph construction process, with red blocks indicating central nodes and blue blocks indicating adjacent nodes. By selecting and clustering central nodes and their adjacent nodes, the model captures discriminative features and manifold structures within the image, thus improving the accuracy of flexible objects recognition.}
\end{figure*}

\subsection{Learning of Channel-Aware Saliency}
%Graph Structure Generation}
Flexible objects exhibit more diverse variations in shape and size, which pose challenges to representation learning. To tackle this, we optimize channel-aware saliency based on attention-driven graph construction, which dynamically captures node relationships by extracting the weight saliency information from each channel, corresponding to the varied shapes and sizes of flexible objects. Furthermore, the model's ability to detect fine-grained features is specifically improved, enabling it to more efficiently distinguish flexible objects from their backgrounds.
 
We divide an image into \begin{math} {N} \end{math} patches and utilize graph embedding to associate them with node vectors \begin{math} V=[v_1,v_2,\ldots,v_N] \in \mathbb{R}^{B \times N \times D} \end{math}, where \begin{math} {B} \end{math} is batchsize and \begin{math} {D} \end{math} denotes feature dimensions. For each node \begin{math} {v_i} \end{math} within the graph, we compute the Euclidean distance to identify the \begin{math} k \end{math} nearest neighbor nodes, collectively denoted as \begin{math} \mathcal{N}(v_i) \in \mathbb{R}^{B \times K \times D} \end{math}. For each neighboring node \begin{math} {v_j} \end{math}, we establish an edge \begin{math} {e}_{ij} \end{math} connecting it to \begin{math} {v}_{i} \end{math}. By traversing all nodes, the set of edges \begin{math} {E} \end{math} is obtained. Consequently, we can construct the graph as \begin{math} G=(V,E) \end{math}.

To improve comprehension of intrinsic nodes relationships of vectors \begin{math} {V}  \in \mathbb{R}^{B \times N \times D} \end{math}, we map these \begin{math} {D} \end{math}-dimensional features to a \begin{math} D^{\prime} \end{math}-dimensional linear latent space via a trainable matrix \begin{math} \mathbf{W} \in \mathbb{R}^{D \times D^{\prime}} \end{math}, and the transformed node vectors \begin{math} V^{\prime} \in \mathbb{R}^{B \times N \times D {\prime}} \end{math} is obtained.
\begin{equation}
V^{\prime}=V\cdot W
\end{equation}
In addition, we define a learnable vector \begin{math} {W}_{s} = [{W}_{s1} \| {W}_{s2}] \in \mathbb{R}^{2D \times 1} \end{math} to project the node feature vectors \begin{math} {V} \end{math} into the attention space to calculate the saliency score, where \begin{math} \| \end{math} denotes as concatenate operation, \begin{math} {W}_{s1} \in \mathbb{R}^{D \times 1} \end{math} means the feature projection matrix of the node itself, and \begin{math} {W}_{s2} \in \mathbb{R}^{D \times 1} \end{math} represents the feature projection matrix of each node with respect to all other nodes. So, we can compute the saliency relationships among nodes:
\begin{equation}
\left\{
\begin{aligned}
S_{s} &= V' \cdot W_{s1} \\
S_{n} &= (V' \cdot W_{s2})^T
\end{aligned}
\right.,
\end{equation}
where \begin{math} {S}_{s} \in \mathbb{R}^{B \times N \times 1} \end{math} denotes saliceny score of the \begin{math} N \end{math} nodes itself, \begin{math} {S}_{n} \in \mathbb{R}^{B \times 1 \times N} \end{math} denotes saliceny score of each central node versus all \begin{math} N \end{math} neighboring nodes. Then, we calculate the attention between nodes saliency score \begin{math} {S}_{s} \end{math} and \begin{math} {S}_{n} \end{math} by broadcast addition, gaining the saliency attention matrix \begin{math} S \in \mathbb{R}^{B \times N \times N} \end{math}:
\begin{equation}
S = \begin{bmatrix}
{S}_{s1} \\
{S}_{s2} \\
\vdots \\
{S}_{sN}
\end{bmatrix} + \begin{bmatrix}
{S}_{n1} & {S}_{n2} & \cdots & {S}_{nN}
\end{bmatrix} = \begin{bmatrix}
{S}_{s1} + {S}_{n1} & {S}_{s1} + {S}_{s2} & \cdots & {S}_{s1} + {S}_{nN} \\
{S}_{s2} + {S}_{n1} & {S}_{s2} + {S}_{n2} & \cdots & {S}_{s2} + {S}_{nN} \\
\vdots & \vdots & \ddots & \vdots \\
{S}_{sN} + {S}_{n1} & {S}_{sN} + {S}_{n2} & \cdots & {S}_{sN} + {S}_{nN}
\end{bmatrix}.
\end{equation}
Since \begin{math} {W}_{s1} \end{math} and \begin{math} {W}_{s2} \end{math} are learnable, saliency attention \begin{math} S \end{math} is capable of adaptively capturing the weight information of the feature channels. Each element \begin{math} {s}_{ij} \end{math} in matrix \begin{math} S \end{math} denotes the attention score between node \begin{math} i \end{math} and \begin{math} j \end{math}. Additionally, we apply normlization methods to obtain the weighted saliceny attention matrix between nodes:
\begin{equation}
\alpha_{ij}=\mathrm{softmax}({s}_{ij})=\frac{\exp(s_{ij})}{\sum_{k\in\mathcal{N}_i}\exp(s_{ik})}.
\end{equation}
Otherwise, \begin{math} \alpha_{ij} \end{math} is transformed by the non-linear LeakyReLU activation function to improve feature diversity, we can obtain:
\begin{equation}
\alpha_{ij} = \frac{\exp(\text{LeakyReLU}(s_{ij}))}{\sum\limits_{k \in \mathcal{N}_i} \exp(\text{LeakyReLU}(s_{ik}))}
\end{equation}

Reviewing the graph construction methodology for ViG, we initially compute the Euclidean distance between all nodes, which forms the node distance matrix \begin{math} {eudist} \end{math}. Subsequently, we apply the \begin{math} k \end{math}-nearest neighbors (KNN) algorithm to determine the \begin{math} k \end{math}-nearest nodes, leading to the creation of the adjacency matrix:
\begin{equation}
Adj=Top-k\left(eudist,k\right),
\end{equation}
using the saliency attention derived from Equation 5, we develop a trainable distance metric \begin{math} \alpha_{ij}*(-eudist) \end{math} that relies on the node relationships \begin{math} {e}_{ij} \end{math}. As a result, we generate an adaptive adjacency matrix based on salicency attention :
\begin{equation}
saliencyAdj=Top-k\left(\alpha_{ij}*eudist,k\right).
\end{equation}

\subsection{Learning of Spatial-Aware Saliency}
%\subsection{Graph Cluster}
Recognizing flexible objects presents a significant challenge due to the minimal differences between classes, which makes them difficult to distinguish. In response, we optimize spatial-aware saliency based on node-level clustering designed to harness underexploited information from nodes within dilated neighboring areas to enhance central nodes. This method promotes interactions within the local context, thus enhancing discrimination capabilities and ensuring better distinction between different categories of flexible objects.

For each central node \begin{math} v_{i,}(i=1,2...,N) \end{math} and its neighboring node \begin{math} v_j\in\mathcal{N}(v_i) \end{math}, in order to fully leverage the dilated information, we employ all \begin{math} k \end{math} neighboring nodes of \begin{math} v_i \end{math} to fuse the node information of \begin{math} \mathcal{N}(v_i) \end{math} using clustering methods; Then, we define a cluster center \begin{math} {c}_{i} \end{math} for \begin{math} \mathcal{N}(v_i) \end{math} by performing average pooling, and node feature similarity \begin{math} s_j \end{math} is obtained between \begin{math} c_i \end{math} and \begin{math} v_j \end{math} based on cosine similarity calculation. Consequently, we obtain the clustered node representation as follows:
\begin{equation}
C_f=\frac{1}{\lambda}\Bigg(c_i+\sum_{j=1}^k\sigma\Big(\alpha s_j+\beta\Big)*\nu_j\Bigg)\\\\\lambda=1+\sum_{j=1}^k\sigma\Big(\alpha s_j+\beta\Big),
\end{equation}
where \begin{math} \alpha \end{math} and \begin{math} \beta \end{math} are learnable parameters that enable \begin{math} s_j \end{math} to adaptively change, and \begin{math} \lambda \end{math} is a regularization term. To maintain adequate representational capabilities, it is essential to incorporate at least one learnable linear transformation to convert the input node features into higher-order features. A 1x1 convolution is employed for feature mapping by Coc \cite{ma2023image}, yet this method notably increases the computational complexity. Consequently, we utilize a learnable linear transformation matrix \begin{math} W\in\mathbb{R}^{D^{'}\times D} \end{math} to facilitate dimension transformation of the initial node feature vectors, thereby diminishing computational overhead. Furthermore, we implement multiple heads, numbered at \begin{math} M \end{math}, to improve the clustering effect. Thus, we obtain:
\begin{equation}
C_f^{\prime}=
\Concat_{m=1}^{M}\frac1\lambda{\left(W\cdot c_i^m+\sum_{j=1}^k\sigma{\left(\alpha s_j^m+\beta\right)*\left(W\cdot\nu_j^m\right)}\right)},
\end{equation}

subsequently, the clustered features \begin{math} C_f^{\prime} \end{math} are adaptively allocated to each node within the cluster according to their similarity. These nodes communicate with each other and share the features among them within the cluster. A linear transformation matrix \begin{math} W^{\prime}\in\mathbb{R}^{D\times D^{\prime}} \end{math} is adopted, and for each node \begin{math} v_j \end{math}, the following update is applied:
\begin{equation}
\nu_j^{\prime}=\nu_j+W^{\prime}\cdot\left(\sigma\left(\alpha s_j+\beta\right)*C_f^{\prime}\right).
\end{equation}

\subsection{Graph Reason Learning}
Previously, we effectively clustered information from adjacent nodes. In this section, we explore the learning interactions between central nodes and their adjacent nodes within the graph. This involves facilitating the learning of node features through processes of aggregation and updating. The process of learning node representations is described as follows:

\begin{equation}
G^{\prime}=Update(Aggregate(G,W_{agg}),W_{update}),
\end{equation}
where \begin{math} G \end{math} is the graph built on the clustered feature \begin{math} C_f^{\prime} \end{math}, \begin{math} {W}_{agg} \end{math} and \begin{math} {W}_{update} \end{math} are the parameters for aggregation and updating, respectively. The graph convolution consists of \begin{math} l \end{math} layers (In this paper, setting \begin{math} l \end{math} = 12.), and the outputs of the upper layers serve as inputs to the lower layers through stacking. Finally, to increase the diversity of features, we utilize a feed-forward neural network (FFN) to map the node features.

\section{FLEXIBLE DATASET}
\begin{minipage}[t]{0.55\textwidth}
\vspace{0pt} % 保证内容从顶部开始
The advancement of discriminative deep learning models is crucially dependent on the access to high-quality datasets. Datasets designed for particular recognition tasks such as fire \cite{jadon2019firenet, foggia2015real} and cloud \cite{krizhevsky2009learning, zhang2018cloudnet} are often narrowly focused, offering limited diversity and challenges. To address these constraints, we propose the FDA—a publicly available, high-quality dataset that includes a wide variety of flexible objects. The FDA are able to be downloaded for research purposes through this \href{https://anonymous.4open.science/r/FDA-D02C/README.md}{access link}.

\end{minipage}%
\hfill %
\begin{minipage}[t]{0.43\textwidth} % Use half of the text width
  \vspace{0pt} % 保证内容从顶部开始
  \centering
  \captionof{table}{Statistics on FireNet Dataset}
\begin{tabular}{ccc}
\toprule % Replaces \hline for a thicker top horizontal line
Categorize & Train & Test \\
\midrule % Aesthetic medium-thickness horizontal line
Fire & 1124 & 593 \\
NoFire & 1301 & 278 \\
total & 2425 & 871 \\
\bottomrule % Replaces \hline for a thicker bottom horizontal line
\end{tabular}
\end{minipage}

\subsection{FireNet Dataset}
Jadon et al. \cite{jadon2019firenet} have constructed a flame dataset for training fire detection models \cite{jadon2019firenet}, comprising 3,296 images. This dataset is divided into two categories: images that contain fire and those without. Table 1 shows the specific details of the image distribution within these categories for both training and test sets. The majority of the FireNet images, sourced online, are solely focused on detecting flames and do not incorporate other variable objects, thus reducing the dataset's realism and diversity. Furthermore, as illustrated in Figure 1 (c), fires are generally captured from a close distance with clear fire characteristics, simplifying the detection process and allowing a lightweight convolutional neural network to achieve accuracies exceeding 90\%.

\subsection{Our FDA Dataset}
Comparable in scale to the FireNet dataset \cite{jadon2019firenet}, our FDA is carefully selected based on the unique characteristics of flexible objects. Specifically, we gathered 2,080 images of flexible objects from real-world scenarios, organizing them into nine fine-grained categories: cloud, facsmoke, fire, fog, glare, ground, light, smoke and water. Additionally, we supplemented this collection with nearly ten thousand images sourced from the Internet, including not only flexible objects but also animals, humans, and vehicles, thus enhancing the diversity of instance. From this extensive collection, we handpicked 394 challenging images to enrich our dataset, thus creating a more diverse and realistic repository. Our flexible objects samples were deliberately chosen for their varying degrees of transparency, shapes, and sizes to highlight their complexity and diversity. Each sample was precisely categorized to ensure accuracy in classification. The final dataset contains 2,474 images,

\begin{minipage}[t]{0.55\textwidth}
\vspace{0pt} % 保证内容从顶部开始
with detailed descriptions of each category presented in Table 2. Figure 1(b) presents several FDA example images, providing an overview of the entire flexible objects dataset. In contrast to the FireNet dataset, our images are primarily sourced from real-world scenarios and range from close-up to distant perspectives. The flexible objects exhibit common characteristics, such as variable sizes, shapes, and semitransparency. However, they also display unique traits that contribute to their complexity. This diversity not only poses significant challenges for the recognition of flexible objects, but also establishes a effective benchmark for evaluating methods of flexible objects recognition.
\end{minipage}%
\hfill %
\begin{minipage}[t]{0.43\textwidth} % Use half of the text width
  \vspace{0pt} % 保证内容从顶部开始
  \centering
  \captionof{table}{Statistics on our FDA}
  \resizebox{0.85\textwidth}{!}{%
  \begin{tabular}{ccc}
\toprule
Categorize & Real-World & Online \\
\midrule
cloud & 201 & 69 \\
facsmoke & 68 & 41 \\
fire & 422 & 39 \\
fog & 262 & 66 \\
glare & 98 & 25 \\
ground & 206 & 42 \\
light & 109 & 25 \\
smoke & 437 & 54 \\
water & 277 & 33 \\
total & 2080 & 394 \\
\bottomrule
\end{tabular}
}
\end{minipage}

\section{Experiments}
In this section, extensive experiments are performed to confirm the challenge of the proposed FDA and the performance of FViG. Comprehensive details of these experiments are included in the Appendix.

\begin{table*}[ht]
\centering
\caption{Comparison results of FViG with current SOTA models on FDA}
\label{tab:architectures}
\resizebox{0.84\textwidth}{!}{%
\begin{tabular}{l|lccc}
\toprule
Category & Method & Parameters(M) & Computation(G) & Accuracy(\%) \\
\midrule
Transformer & ViT-B/16 \cite{dosovitskiy2020image} & 86.8 & 17.6 & 76.38 \\
 & T2T-ViT-14 \cite{yuan2021tokens} & 21.5 & 4.8 & 75.46 \\
 & DaViT-Small \cite{ding2022davit} & 49.7 & 8.8 & 77.19 \\
 & DeiT-S \cite{touvron2021training} & 22.1 & 4.6 & 78.82 \\
 & Swin-S \cite{liu2021swin} & 50 & 8.7 & 78.58 \\
 & Twins-SVT-S \cite{chu2021twins} & 24 & 2.9 & 30.32 \\
\midrule
CNN & Resnet50 \cite{he2016deep} & 25.6 & 4.1 & 70.37 \\
 & Resnet101 \cite{he2016deep} & 45 & 7.9 & 72.18 \\
 & Regnet \cite{radosavovic2020designing} & 4.78 & 0.406 & 74.65 \\
 & Densenet \cite{huang2017densely} & 18 & 4.37 & 71.29 \\
\midrule
MLP & Mlp-mixer-base \cite{tolstikhin2021mlpmixer} & 59 & 12.7 & 68.76 \\
 & Mlp-mixer-larger \cite{tolstikhin2021mlpmixer} & 207 & 44.8 & 68.17 \\
\midrule
Graph & Vig-s \cite{han2022vision} & 22.7 & 4.5 & 74.68 \\
 & Vig-ti \cite{han2022vision} & 7.1 & 1.3 & 75.37 \\
 & Coc-tiny \cite{ma2023image} & 5.3 & 1 & 77.08 \\
 & Coc-small \cite{ma2023image} & 14 & 2.6 & 81.41 \\
 & \textbf{FViG} & \textbf{21.16} & \textbf{4.35} & \textbf{80.72} \\
 & \textbf{FViG-tiny} & \textbf{5.98} & \textbf{1.23} & \textbf{78.10} \\
\bottomrule
\end{tabular}%
}
\end{table*}

\begin{figure*}[h]
  \centering
  \includegraphics[width=0.90\linewidth]{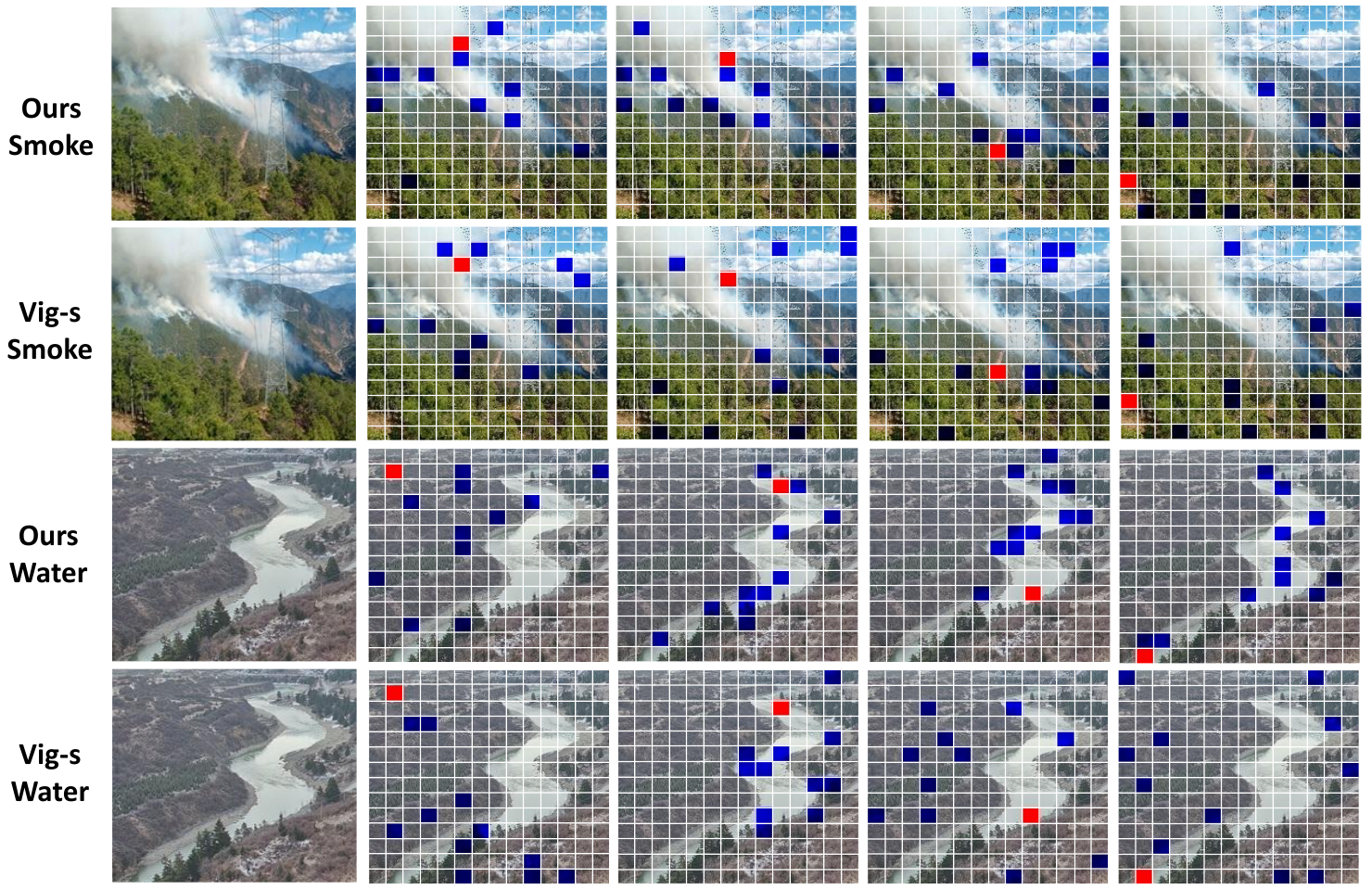}
  \caption{Visualization of the constructed graph structure. For the images of smoke and water, we selected two central nodes from both the foreground and background. The patches represented by these chosen nodes are marked in red, and the nodes that eventually form neighboring relationships with them are marked in blue.}
\end{figure*}

\subsection{Comparison with SOTA Models}
We conducted a comprehensive experimental evaluation on the FDA dataset to substantiate the improved performance of our FViG compared to current SOTA models. As detailed in Table 3, these comparative experiments were rigorously designed to assess the efficacy of the proposed FViG against an array of established models, including those based on Transformers, Convolutional Neural Networks (CNNs), Multilayer Perceptrons (MLPs) and various GNNs. Our FViG and its compact variant, FViG-tiny, showcased in bold, exhibit an impressive balance of parameter efficiency, computational cost, and predictive accuracy. Remarkably, FViG achieves an accuracy of 80.72\%, which is significantly better than traditional CNN and MLP, as well as the other GNN models. Impressively, it achieves this superior accuracy with a reduced parameter count of 21.16 million, which is less than quarter compared to certain Transformer models such as ViT-B/16. Additionally, the computational cost of FViG, at 4.35 Gflops, is relatively low, demonstrating its efficiency, particularly when compared with the MLP-mixer-large model that requires a significant computational overhead of 44.8 Gflops. The tiny version of FViG further emphasizes efficiency, requiring only 5.98 million parameters and 1.23 Gflops, while still achieving a commendable accuracy of 78.10\%. Furthermore, the detailed experiment conducted on the FireNet dataset can be found in Appendix A.3.

\begin{minipage}[t]{0.48\textwidth}
\vspace{0pt} % 保证内容从顶部开始
  %\centering
  \vspace{0pt} % 保证内容从顶部开始
   This variant offers a viable alternative for scenarios with constrained computational resources that maintain high performance without considerable sacrifice. Our models, which stand out by their innovative architectural design, excel not only in diminishing computational demands but also in augmenting the accuracy of flexible objects recognition. This highlights the effectiveness of our method in addressing the complexities and challenges associated with flexible objects recognition.
\end{minipage}%
\hfill %
\begin{minipage}[t]{0.50\textwidth} % Use half of the text width
  \centering
  \vspace{0pt} % 保证内容从顶部开始
\captionof{table}{Ablation results of our FViG with various configurations.}
\label{tab:methods_accuracy}
\resizebox{\textwidth}{!}{%
\begin{tabular}{lcccc}
\toprule
Baseline & Channel & Spatial & Dilation & Accuracy(\%) \\
         & Saliency & Saliency & & \\
\midrule
\checkmark &  &  &  & 74.68 \\
\checkmark & \checkmark &  &  & 77.3(\textcolor{green}{\textcolor{red}{$\uparrow$}2.62})\\
\checkmark &  & \checkmark &  & 78.9(\textcolor{green}{\textcolor{red}{$\uparrow$}4.22}) \\
\checkmark & \checkmark & \checkmark &  & 79.13(\textcolor{green}{\textcolor{red}{$\uparrow$}4.45}) \\
\checkmark &  & \checkmark & \checkmark & 79.01(\textcolor{green}{\textcolor{red}{$\uparrow$}4.33}) \\
\checkmark & \checkmark & \checkmark & \checkmark & 80.72(\textcolor{green}{\textcolor{red}{$\uparrow$}6.04}) \\
\bottomrule
\end{tabular}
}
\end{minipage}

\subsection{Ablation Study}
We analyze each part of the model to determine their individual contributions to the overall performance. Table 4 in our study presents various configurations of our method along with their corresponding accuracy. The baseline ViG achieved an accuracy of 74.68\%. The introduction of channel-aware saliency module resulted in a 2.62\% increase in accuracy, indicating that saliency attention could improve the model sensitivity to distinguish flexible objects. The introduction of the spatial-aware saliency module resulted in a 4.22\% increase in accuracy, indicating that the Cluster is able to facilitate interaction in the local context and improve discrimination ability. The introduction of channel-aware and spatial-aware saliency modules led to a substantial performance increase, with an accuracy of 79.13\%. The adoption of the spatial-aware saliency and Dilation modules resulted in a 4.33\% increase in accuracy, indicating that the Dilation module could enhance the feature of central nodes by establishing graph relationships between nodes within dilated regions. Ultimately, when all modules were incorporated, our method achieved an accuracy of 80.72\%. 

\subsection{Visualization}
To better understand the workings of our FViG, we visualized the graph structure constructed within the FViG and compared it with the ViG model. In Figure 3, we illustrate the differences in graph structure when using the FViG and ViG models for two different categories of input samples (smoke and water). For the images of smoke and water, we selected two central nodes in both the foreground and background, respectively. The patches corresponding to these nodes are colored red, whereas the nodes that subsequently establish neighboring relationships are colored blue. We observed that when the patch of a chosen central node represents the foreground, the FViG tends to select neighboring nodes primarily from the foreground. Conversely, when the central node's patch represents the background, the neighboring nodes chosen are mostly from the background. But ViG's performance is much worse, whether the central node is foreground or background. Our model is more capable of selecting neighboring nodes relevant to the current node's content and is more effective in recognizing flexible objects.

\begin{figure*}[h]
  \centering
  \includegraphics[width=0.68\linewidth]{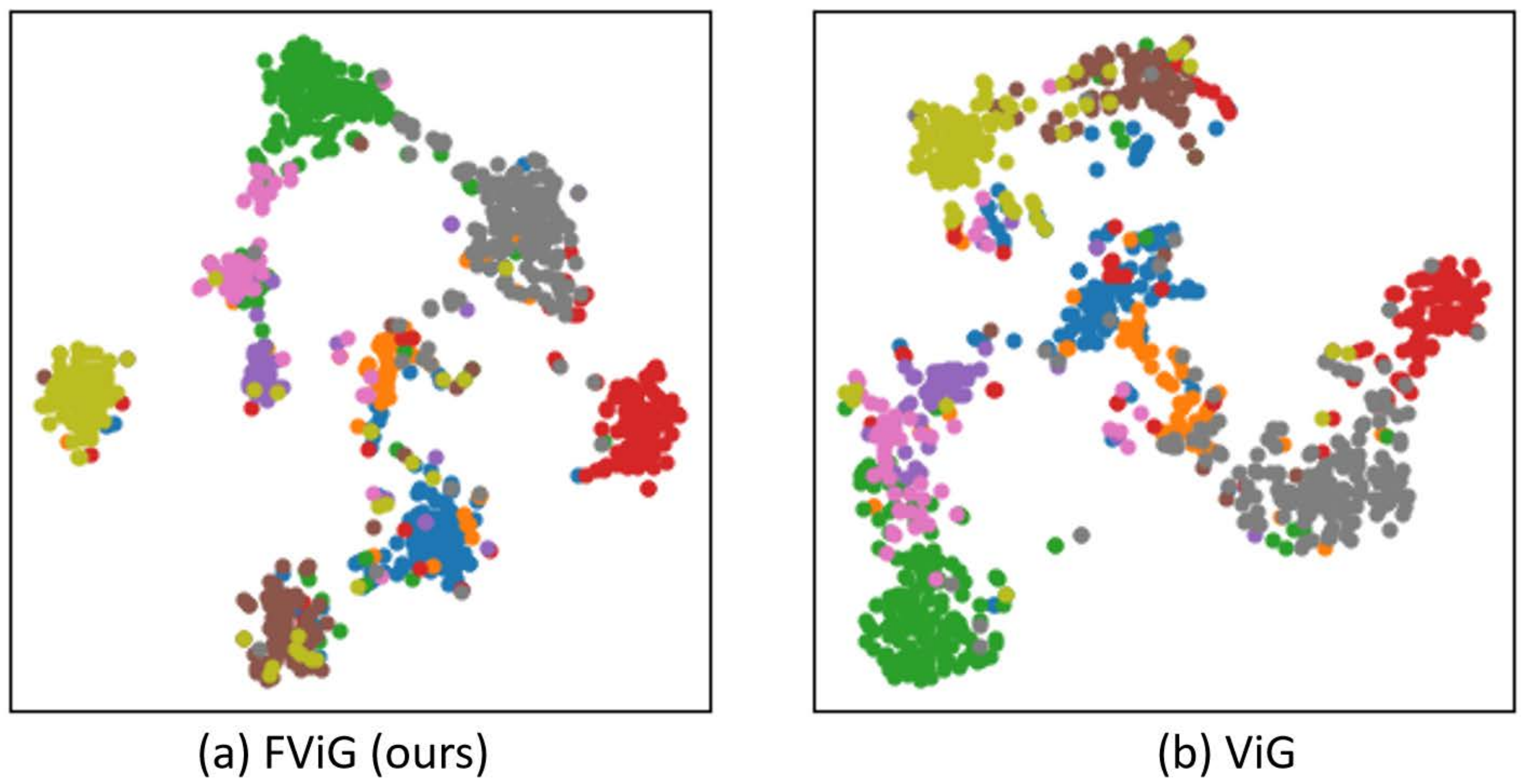}
  \caption{A comparative study of t-SNE visualizations is conducted for our FViG and ViG.}
\end{figure*}

\subsection{t-SNE Analysis}
As illustrated in Figure 4 (a) and Figure 4 (b), we present the t-SNE visualization for FViG and ViG, respectively. In these visualizations, the clustering achieved by FViG appears more cohesive, with clearer demarcations among categories. The vividly colored clusters are well-separated, demonstrating the effectiveness of our method in distinguishing these data points. Additionally, FViG shows fewer outliers, suggesting a higher resilience in handling noise or outliers. In contrast, the clusters in ViG are less distinct, with some colored clusters (such as red and blue) positioned closely, and certain areas exhibiting blurred boundaries. Moreover, ViG displays a greater number of outliers and more dispersed clusters, indicating a potential weakness in dealing with datasets that have indistinct boundaries.

\section{Conclusion}
In this paper, we have constructed a diverse dataset and proposed the Flexible Vision Graph Network (FViG) for the recognition of flexible objects. We address the major challenges in flexible objects recognition by employing channel-aware saliency learning to enable the adaptation of graph representation and spatial-aware saliency learning to improve the discriminative capabilities, respectively. Extensive experiments on the FDA and FireNet datasets demonstrate the effectiveness of our proposed FViG, achieving recognition performance comparable to other SOTA methods. Our study indicates the efficacy of FViG in fine-grained and irregular features of flexible objects. Despite its advantages, while FViG performs well on datasets specifically curated for flexible objects, its generalizability to other domains or more diverse datasets remains to be thoroughly investigated.

\bibliographystyle{unsrt}
\bibliography{flexible}

%%%%%%%%%%%%%%%%%%%%%%%%%%%%%%%%%%%%%%%%%%%%%%%%%%%%%%%%%%%%

\appendix

\section{Appendix}
\subsection{Experimental Settings}
In the experimental configuration of our FViG, we fine-tuned various hyperparameters to enhance performance. The initial learning rate for the FViG was established at 0.3125e-4, based on the learning rate of 2e-3 from the ViG model, and tailored to fit our specific hardware configuration and training approach. While the ViG model was run using 8 GPUs and a batch size of 128, our experiments utilized a single GPU with a batch size of 16. As a result, we calculated our learning rate by dividing the ViG's rate of 2e-3 by 64 (reflecting the 8 GPUs and the 8 times smaller batch size), to maintain stable training given our more constrained hardware capabilities. Our model completed 100 epochs of training using the AdamW optimizer.  

\begin{minipage}[t]{0.55\textwidth}
\vspace{0pt} % 保证内容从顶部开始
The learning rate followed a cosine schedule, known for promoting efficient convergence. To mitigate overfitting, we applied a dropout rate of 0.1. The architecture of the model featured 12 adjacent nodes and adopted a dynamic dilation rate that escalated with each layer's depth, increasing by 1 every 4 layers. This method of adaptive dilation allowed the model to perceive a wider range of spatial relationships. Furthermore, the clustering module incorporated 4 multi-heads, improving the model's capability to synthesize detailed features. Details of these configurations are provided in Table 5. The networks were implemented using PyTorch and trained on a single NVIDIA RTX 3090 GPU.
\end{minipage}%
\hfill %
\begin{minipage}[t]{0.43\textwidth} % Use half of the text width
  \vspace{0pt} % 保证内容从顶部开始
  \centering
  \captionof{table}{Hyperparameters for FViG}
  \resizebox{0.87\textwidth}{!}{%
\begin{tabular}{lc}
\toprule
Hyper-parameters & Value \\
\midrule
Batch size & 16 \\
Learning rate & 2e-3/64 \\
Epochs & 100 \\
Optimizer & AdamW \\
Learning rate schedule & Cosine \\
Dropout rate & 0.1 \\
Adjacent nodes & 12 \\
Dilation rate & 1, 2, 3, 4 \\
Multi-heads & 4 \\
\bottomrule
\end{tabular}
}
\end{minipage}

\begin{minipage}[t]{0.50\textwidth} % Use half of the text width
  \centering
  \vspace{0pt} % 保证内容从顶部开始
  \includegraphics[width=\linewidth]{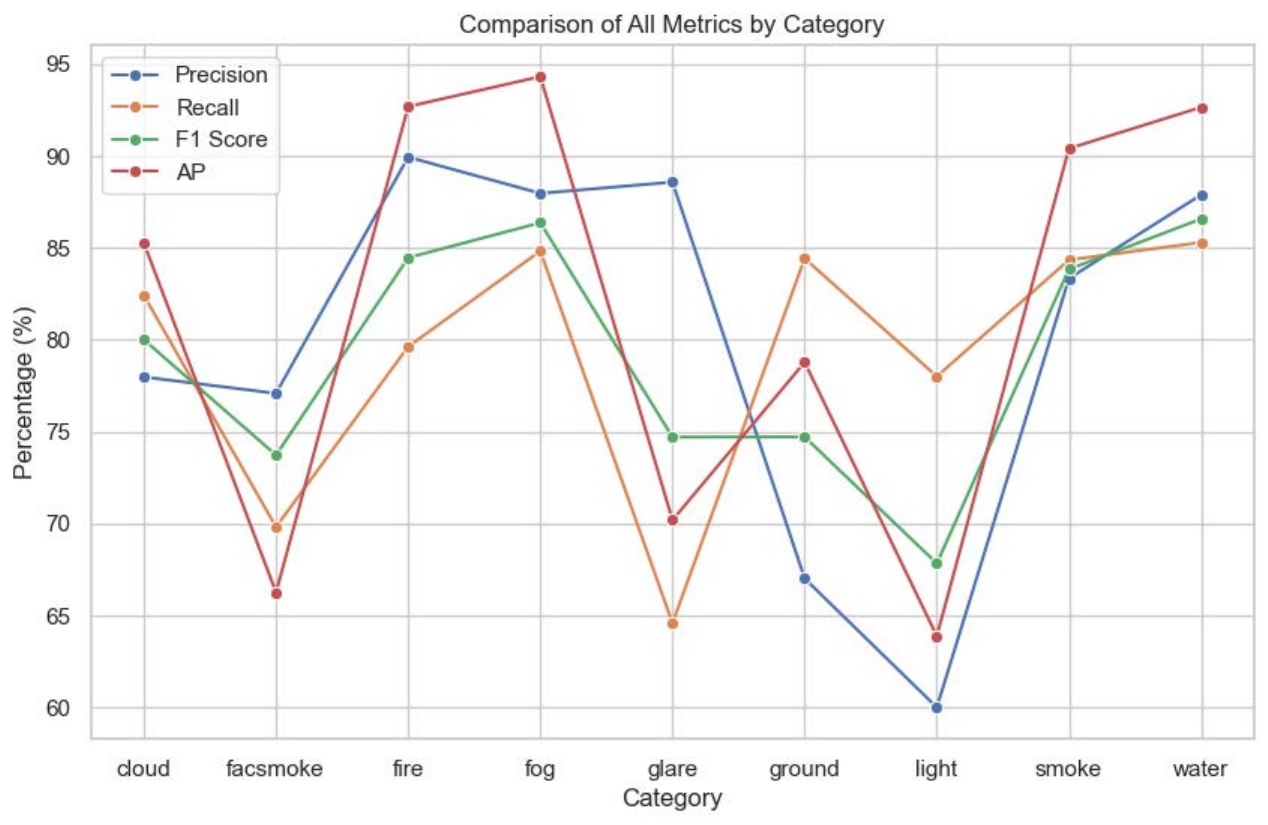}
  \captionof{figure}{Classification performance metrics of nine categories in the dataset.}
\end{minipage}
\hfill %
\begin{minipage}[t]{0.48\textwidth} % Use half of the text width
  \centering
  \vspace{0pt} % 保证内容从顶部开始
  \includegraphics[width=0.76\linewidth]{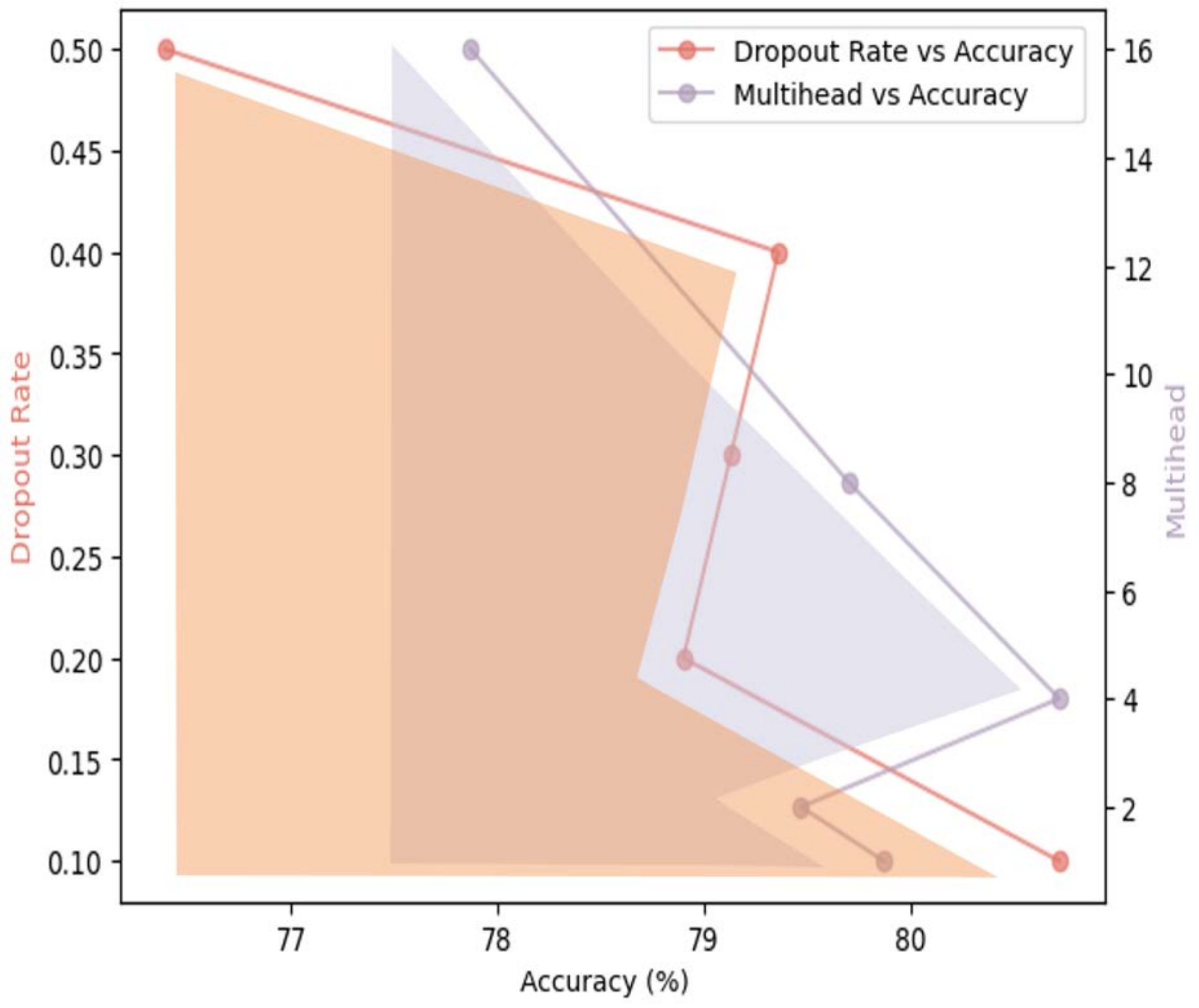}
  \captionof{figure}{Accuracy of various dropout rate and multihead numbers.}
\end{minipage}

\subsection{Classification Results on FDA}
In this section, we discuss the classification results obtained in FDA, focusing on performance within nine distinct object categories. For a complete assessment, we examine several metrics per category, such as Precision, Recall, F1 Score, and Average Precision. Additionally, we employ Precision-Recall (P-R) curves and Receiver Operating Characteristic (ROC) curves to visually depict our model's performance, and we also present confusion matrices to offer an in-depth analysis of the classification precision for each category of flexible objects.

\begin{minipage}[t]{0.49\textwidth}
\vspace{0pt} % 保证内容从顶部开始
  \centering
  \vspace{0pt} % 保证内容从顶部开始
  \includegraphics[width=\linewidth]{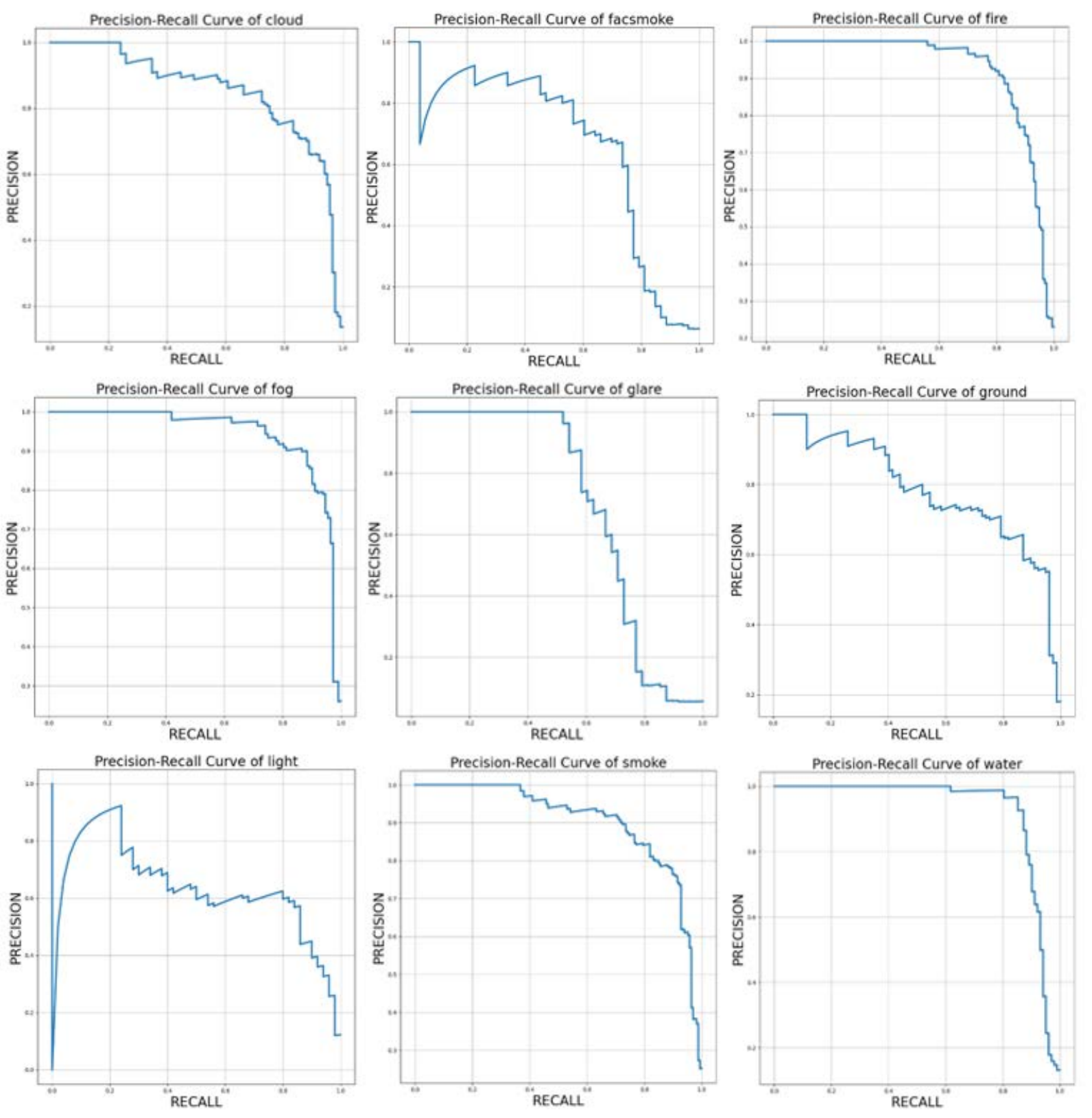}
  \captionof{figure}{Precision-Recall curves for the classification of nine categories.}
\end{minipage}%
\hfill %
\begin{minipage}[t]{0.48\textwidth} % Use half of the text width
  \centering
  \vspace{0pt} % 保证内容从顶部开始
  \includegraphics[width=\linewidth]{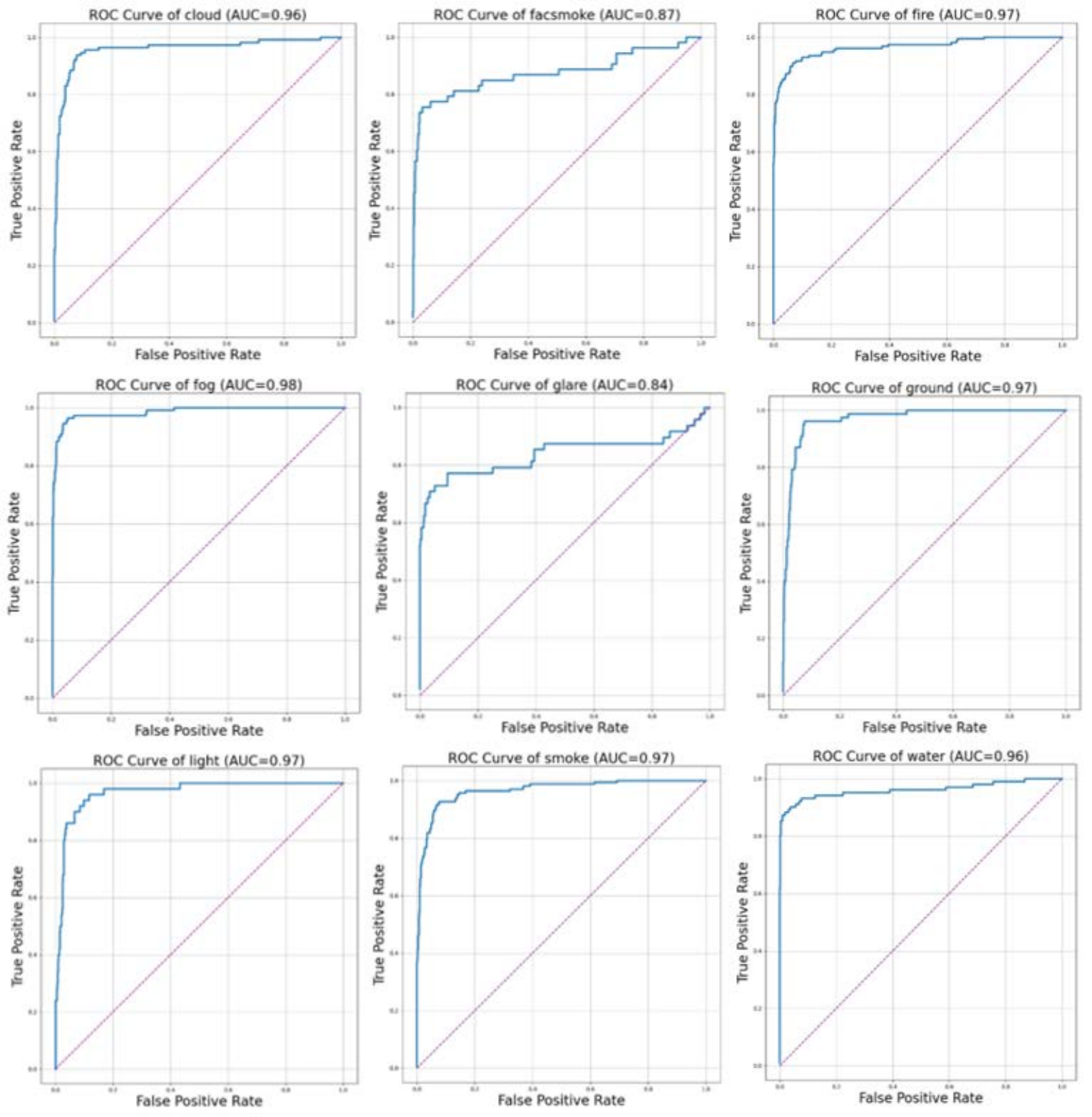}
  \captionof{figure}{ROC curves for the classification of nine categories.}
\end{minipage}

Figure 5 shows that the model achieves high precision, recall, F1 scores, and AP for categories such as cloud, fire, fog, smoke, and water, demonstrating its strong recognition capabilities. However, categories such as ground and light exhibit relatively lower metrics, with precision rates at 67.01\% and 60.00\%, respectively. Furthermore, the recall rates for facsmoke and glare are 69.81\% and 64.58\%, respectively. These lower scores may be attributed to the less distinct features of these flexible objects and the higher rate of misclassification caused by small differences between classes. Nevertheless, the overall performance of the model across all categories remains solid, affirming the effectiveness of the proposed FViG model in recognizing flexible objects of various shapes and sizes.

\begin{minipage}[t]{0.50\textwidth}
\vspace{0pt} % 保证内容从顶部开始
  %\centering
  \vspace{0pt} % 保证内容从顶部开始
  \includegraphics[width=0.9\linewidth]{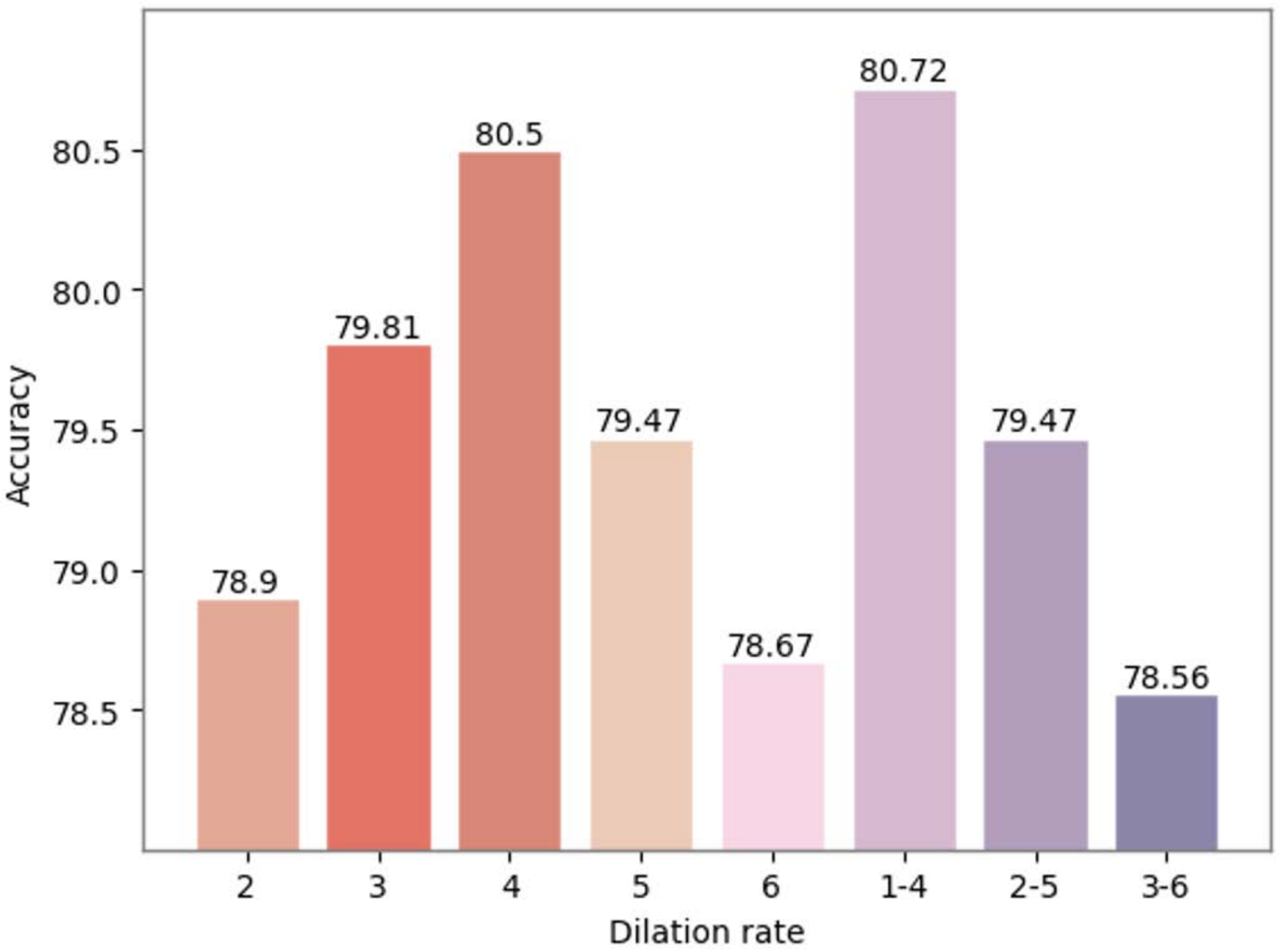}
  \captionof{figure}{Accuracy of various dilation rate.}
\end{minipage}%
\hfill %
\begin{minipage}[t]{0.48\textwidth} % Use half of the text width
  \centering
  \vspace{0pt} % 保证内容从顶部开始
  \includegraphics[width=0.9\linewidth]{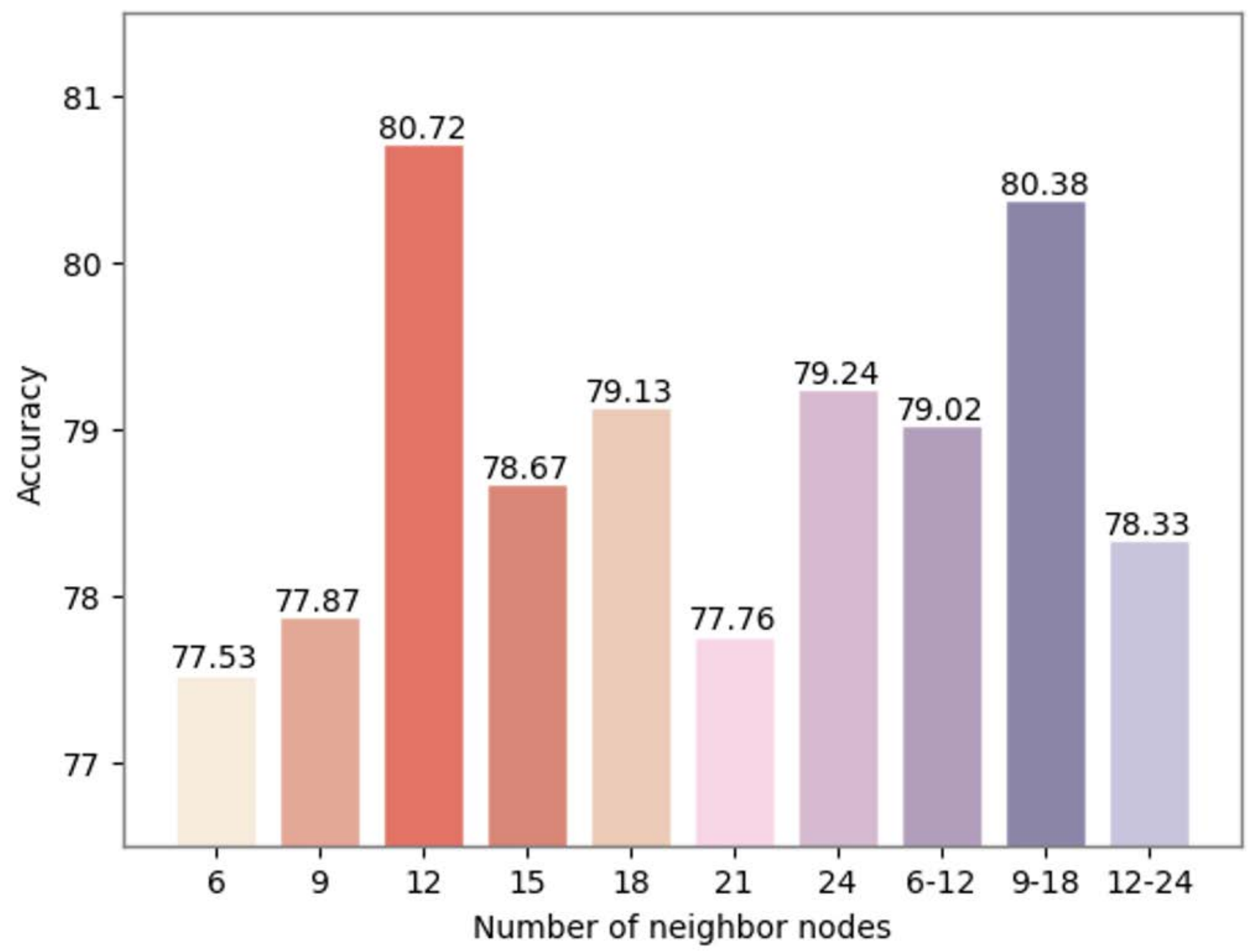}
  \captionof{figure}{Accuracy of various number of adjacent nodes.}
\end{minipage}

The P-R curve facilitates an understanding of the trade-off between precision and recall at various thresholds. Figure 7 shows the P-R curves for the nine category of objects classified by the FViG. Similarly, the ROC curve demonstrates the relationship between the true positive rate and the false positive rate at various thresholds. The area under the ROC curve (AUC) serves as a comprehensive summary of the model's performance, with a value of 1 indicating perfect accuracy and a value of 0.5 suggesting no discriminative ability. Figure 8 presents the ROC curves for the nine categories of objects recognized by the FViG. Through the analysis of both P-R and ROC curves, the optimal threshold that achieves a balance between precision and recall can be identified, adapted to the specific needs of the application.

The confusion matrix for the FViG classification in the FDA is depicted in Figure 11. This matrix highlights the performance of the FViG approach in classifying nine categories of flexible objects. Notably, the matrices show remarkable accuracy in recognizing fire and smoke, demonstrating the model's capability in distinguishing detail features that frequently confuse human perception. Additionally, the model also achieved commendable performance in other categories.

\subsection{Comparison of FDA and FireNet datasets}
Furthermore, experiments were carried out on the FireNet dataset. As indicated in Table 6, we compared several SOTA techniques, with our FViG consistently outperforming others. However, a trend observed in the experimental results is the unusually high accuracy rates for all SOTA methods tested on the FireNet dataset. This consistent pattern implies that the FireNet dataset lacks the necessary complexity and diversity to adequately test and assess sophisticated object recognition models. Such uniformly excellent results suggest that the dataset fails to accurately reflect the complexity and unpredictability of real-world scenarios, particularly in terms of flexible objects recognition. In comparison, our FDA dataset poses a greater challenge and is more appropriate to push forward research in flexible objects recognition. It encompasses a broader range of scenarios and object types, adding complexities that better replicate the challenges found in real-world settings. 

\begin{minipage}[t]{0.48\textwidth}
\vspace{0pt} % 保证内容从顶部开始
%\centering
\captionof{table}{Comparison of SOTA Methods and FViG on FireNet Dataset}
\label{tab:methods_accuracy}
\renewcommand{\arraystretch}{1.30}
\resizebox{\textwidth}{!}{%
\begin{tabular}{lcc}
\toprule
Category & Method & Accuracy(\%) \\
\midrule
Transformer & ViT-B/16 \cite{dosovitskiy2020image} & 89.81 \\
                             & Swin-s \cite{liu2021swin} & 94.32 \\
                             & T2T-ViT-14 \cite{yuan2021tokens} & 93.63 \\
CNN & Resnet50 \cite{he2016deep} & 90.05 \\
                     & Regnet \cite{radosavovic2020designing} & 95.94 \\
MLP & Mlp-mixer-base \cite{tolstikhin2021mlpmixer} & 93.03 \\
Graph & Coc-small \cite{ma2023image} & 98.73 \\
                       & \textbf{FViG} & \textbf{98.85} \\
\bottomrule
\end{tabular}
}
This diversity not only evaluates the performance and adaptability of object recognition algorithms but also fosters the development of more sophisticated and nuanced models capable of handling the intricacies of real-world data. Thus, our FDA dataset serves as a valuable and challenging benchmark that advances and promotes the field of flexible objects recognition.
\end{minipage}%
\hfill %
\begin{minipage}[t]{0.50\textwidth} % Use half of the text width
  \centering
  \vspace{0pt} % 保证内容从顶部开始
  \includegraphics[width=\linewidth]{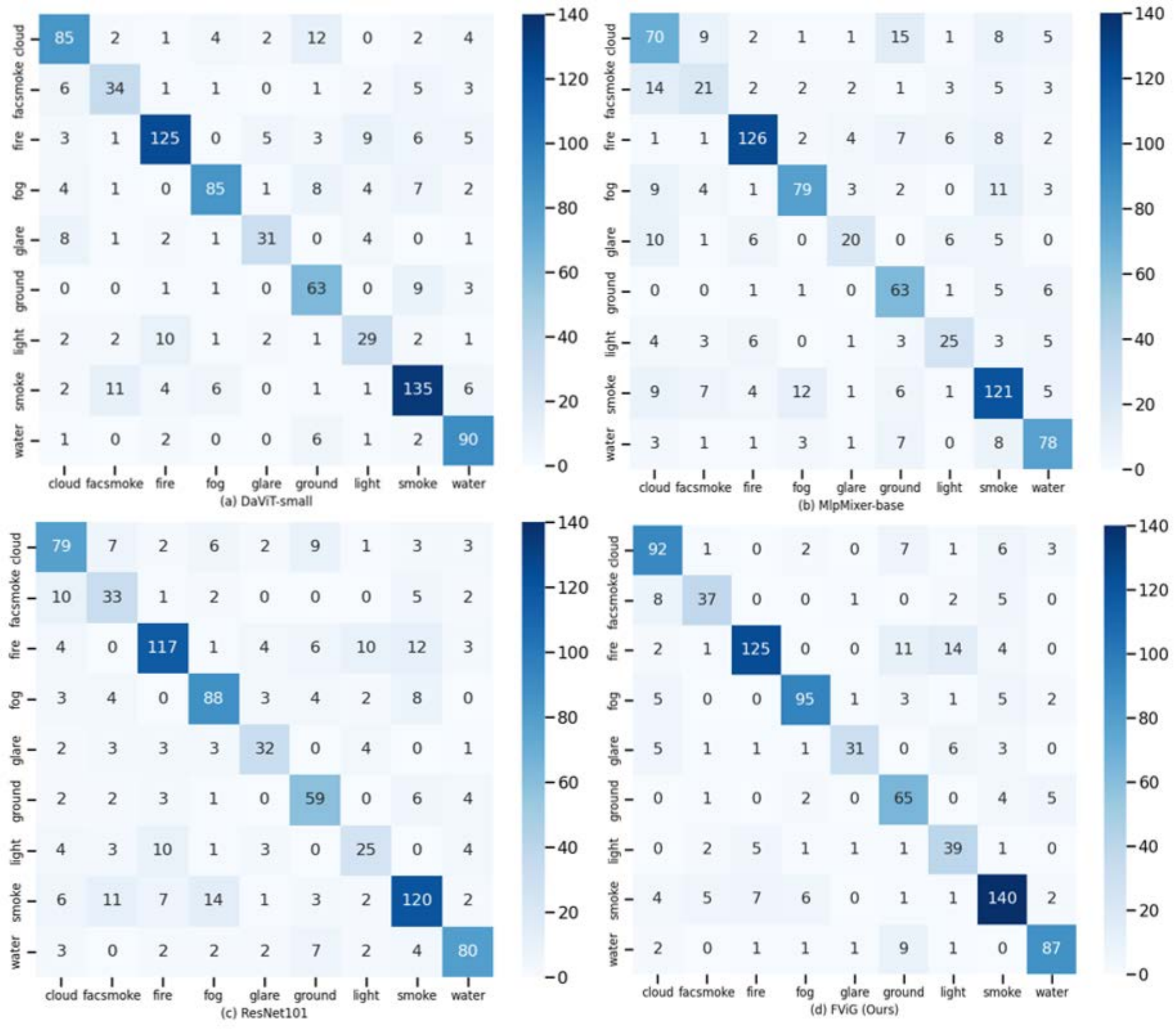}
  \captionof{figure}{The confusion matrix exhibits the classification performance across various categories. These matrices were derived from four models: a) DaViT-small, b) MlpMixer-base, c) ResNet101, d) FViG.}
\end{minipage}

\subsection{Sensitivity Analysis}
Furthermore, we carried out experimental investigations to assess the influence of several hyperparameters on the FViG's performance. These hyperparameters encompass the choice of adjacent node count, the adjustment of the neighboring number, the dilation rate, the dropout ratio, and the multihead count. The results of these experiments are presented in Figure 6, 9 and 10. Based on these results, we ultimately selected a configuration with 12 neighboring nodes (K=12), a dilation rate ranging from 2 to 5 (D = range (2-5), a dropout ratio of 0-0.1, and a multihead count of 4 as our final hyperparameters. With these settings, our FViG achieved its highest accuracy of 80.72\%.

%%%%%%%%%%%%%%%%%%%%%%%%%%%%%%%%%%%%%%%%%%%%%%%%%%%%%%%%%%%%

\end{document}